\documentclass[letterpaper, 10 pt, conference]{ieeeconf}  

\IEEEoverridecommandlockouts                              

\usepackage{multicol}
\usepackage[bookmarks=true]{hyperref}
\usepackage{amsmath}
\usepackage{graphicx}
\usepackage{grffile}
\usepackage{blindtext}
\usepackage[bottom]{footmisc}
\usepackage{mathrsfs}
\usepackage{lipsum}
\usepackage{url}
\usepackage{array}
\usepackage{booktabs}
\usepackage{float}
\usepackage{wrapfig}

\newlength\fwidth
\usepackage{graphicx}
\usepackage{framed}
\usepackage{indentfirst}
\usepackage{multirow}
\usepackage{makecell}

\usepackage{amsmath, amssymb,amsthm}
\usepackage{booktabs}
\usepackage{tabularx}
\usepackage[capitalise]{cleveref}
\usepackage{comment}
\usepackage{color}
\usepackage{tabularx}
\usepackage{algorithmic}
\usepackage{algorithm}
\usepackage{bm}
\usepackage{xcolor}
\usepackage{caption}
\usepackage{subcaption}

\title{\LARGE \bf
Zero-Shot Policy Transfer with Disentangled Task Representation of Meta-Reinforcement Learning
}

\author{Zheng Wu$^{1, *}$, Yichen Xie$^{1, *}$, Wenzhao Lian$^{2}$, Changhao Wang$^{1}$, Yanjiang Guo$^{3}$, \\
Jianyu Chen$^{3}$, Stefan Schaal$^{2}$ and Masayoshi Tomizuka$^{1}$
\thanks{* The authors are equally contributed. }
\thanks{$^{1}$University of California, Berkeley, Berkeley, CA, USA. $^{2}$Intrinsic Innovation LLC, Mountain View, CA, USA. $^{3}$ Tsinghua University, Beijing, China. }%
}

\begin{document}

\maketitle
\thispagestyle{empty}
\pagestyle{empty}

\begin{abstract}
Humans are capable of abstracting various tasks as different combinations of multiple attributes. 
This perspective of compositionality is vital for human rapid learning and adaption since previous experiences from related tasks can be combined to generalize across novel compositional settings. 
In this work, we aim to achieve zero-shot policy generalization of Reinforcement Learning (RL) agents by leveraging the task compositionality. 
Our proposed method is a meta-RL algorithm with disentangled task representation, explicitly encoding different aspects of the tasks.
Policy generalization is then performed by inferring unseen compositional task representations via the obtained disentanglement without extra exploration. 
The evaluation is conducted on three simulated tasks and a challenging real-world robotic insertion task. 
Experimental results demonstrate that our proposed method achieves policy generalization to unseen compositional tasks in a zero-shot manner.
\end{abstract}

\section{INTRODUCTION}
\label{sec:inrto}
\begin{figure}
    \centering
    \includegraphics[width=0.46\textwidth]{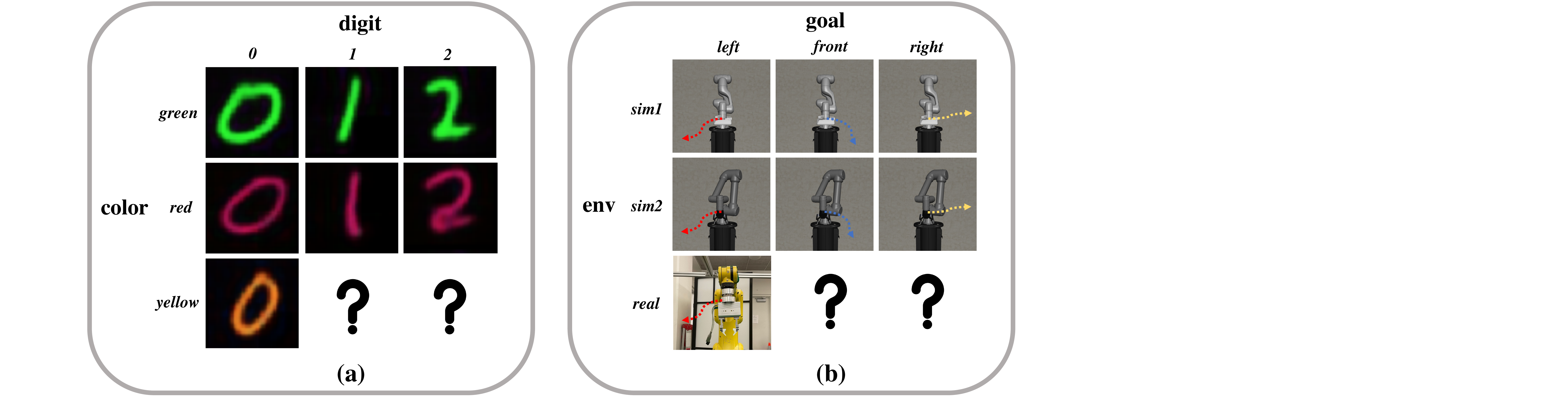}
    \caption{Illustration of unseen compositional tasks with existing degrees of variation. (a) While (\textit{yellow, 1}) image is unseen, both \textit{yellow} and \textit{1} are present in other images. (b) While (\textit{real, front}) is a novel task, both \textit{real} and \textit{front} are present in other tasks.}
    \label{fig:motivation}
\end{figure}

Policy transfer has been a long-standing challenge in the robotics community. 
Reasoning over existing task experiences and transferring knowledge to novel combinatorial tasks with familiar elements is vital for human rapid learning and adaption.
For instance, the images in Figure~\ref{fig:motivation}(a) can be abstracted as combinations of (\textit{digit}, \textit{color}), and humans can effortlessly imagine other unseen combinatorial images. 
In this work, we aim to enable similar ability in the context of reinforcement learning (RL). 
Specifically, we consider the generalization scenarios where tasks-of-interest can be described by a set of degrees of variation (DoVs) and our goal is to achieve policy transfer across unseen compositional tasks, namely, tasks of unseen combinations of existing DoVs. 
Considering the example depicted in Figure~\ref{fig:motivation}(b), the robot policy is influenced not only by the \textit{goal} that can be directly measured, but also by other \textit{environment}-related properties of the system, such as robot model, friction, control gain, etc. 
When facing an unseen compositional task, \textit{e.g., } (\textit{real}, \textit{front}) in Figure~\ref{fig:motivation}(b), we aim to combine the  knowledge obtained from other related tasks, \textit{e.g.,} (\textit{real}, \textit{left}), (\textit{sim1}, \textit{front}), to directly get the policy for the new task.

The idea of policy transfer across novel task combinations of known DoVs is also explored in~\cite{devin2017learning}, where the authors proposed a modular policy architecture to decompose the policy into goal-specific and robot-specific modules and demonstrated zero-shot policy transfer to unseen robot-goal combinations using existing modules. 
Specifically, the authors decompose the observations into goal-related and robot-related observations, which are fed as input to the corresponding neural network modules to output desired actions. 
However, this method requires training an individual neural network based module for each possible value of DoV, making it computationally expensive to scale up when certain DoV have many possible values. 
For example, given 20 different possible goals and 10 different possible robots, this method needs to train 30 different modules in total. 
Additionally, one implicit assumption of~\cite{devin2017learning} is that observation alone can capture the differences between tasks with different DoV combinations. However, there exist certain scenarios where the available observations fail to reflect the differences between tasks. 
For example, when learning an RL agent to push objects of different weights using image as observation, the dynamics variations between tasks are not captured by observation $o_t$, but instead by observed transitions $(o_t, a_t, o_{t+1}, r_t)$. Solely performing task decomposition \textit{w.r.t.} the observation space limits the potential to be applied in such scenarios.

In contrast, the key insight of our work is to achieve task decomposition from the perspective of task representation instead of policy architecture as in~\cite{devin2017learning}. 
Specifically, we build upon the framework of context-based meta-RL framework~\cite{finn2018probabilistic, rakelly2019efficient, zhao2020meld, humplik2019meta, zintgraf2019varibad} which jointly learns a latent task embedding from observed transitions and a policy conditioned on the inferred task representation. 
The task decomposition is achieved by disentangling latent task embedding as the concatenation of embeddings for each task DoV. 
For this purpose, we develop a task disentanglement regularization objective for meta-training (Section~\ref{sec:method:training}). 
The achieved disentanglement in the latent task space allows us to infer novel task representations without extra experiences by combining the existing representations of task DoVs in the unseen tasks, thus achieving zero-shot policy generalization (Section~\ref{sec:method:generalization}).


To evaluate our approach, we extend three traditional meta-RL tasks to compositional tasks composed of certain DoVs and test the generalization performance of our method on the unseen compositional tasks. 
Experimental results indicate that the generalized policies obtained from our method in a zero-shot manner outperform other baselines. 
We also demonstrate that our proposed approach can be seamlessly applied as a zero-shot sim-to-real generalization method. We evaluate the sim-to-real algorithm on challenging real-world tilted peg-in-hole tasks. Results demonstrate the effectiveness of our approach.

\section{Related Work}
\subsection{RL Policy Generalization across Similar Tasks} 
Policy generalization has been a long-standing challenge in robotics community. 
While RL~\cite{sutton2018reinforcement} achieves promising results on robotic manipulation tasks~\cite{kalashnikov2018scalable, lee2019making}, 
the impedance of deploying RL algorithms lies at the numerous data required during training. To this end, many efforts have been devoted to generalizing the existing policies to similar tasks. A common application of policy generalization is sim-to-real~\cite{zhao2020sim}, where a policy is  learned in simulation and then transferred to the real world. This is usually achieved by domain randomization, which randomizes the tasks with a wide task distribution in simulation and assumes the real-world task is captured by this distribution~\cite{tobin2017domain, peng2018sim, pinto2017asymmetric}. 
Meta-RL approaches~\cite{rakelly2019efficient, zhao2020meld, zintgraf2019varibad, duan2016rl, finn2017model, gupta2018meta, zintgraf2019fast} offer another perspective of policy generalization by learning to learn from a distribution of tasks, enabling the agent to quickly adapt to novel tasks with limited explorations.
Context-based meta-RL methods~\cite{rakelly2019efficient, zhao2020meld}, which view meta-RL as task inference and learn a hidden task variable from collected experience, have demonstrated the ability to adapt meta-policy to real-world manipulation tasks~\cite{zhao2020meld, schoettler2020meta}. 
However, those methods still require collecting online rollout data for task inference and are therefore not zero-shot. 
Zero-shot policy transfer is particularly useful in circumstances when online interactions are expensive; for example, the task objects are fragile such as the thru-hole components in PCB assembly.
In this work, we improve context-based meta-RL methods for policy transfer in a zero-shot manner by considering the task decomposition in the latent task representation. 

\subsection{Task Decomposition in Robotics}
Task decomposition is widely investigated in the robotics field to ease the learning process and reuse the knowledge from similar tasks. 
Most of the research efforts focus on decomposing long-horizon manipulation tasks temporally into several sub-tasks and learning sub-policy for each sub-task~\cite{andreas2017modular, haarnoja2018composable, huang2019neural, jiang2019language, kipf2019compile, zhou2022policy}. 
The learned sub-policies that can be re-arranged in a novel way to obtain the policy for unseen long-horizon tasks. 
There are also some works that consider task decomposition from other perspectives. 
For example, ~\cite{oh2017zero} performs task decomposition by leveraging the correspondence in the provided analogy and decompose a task as a (action, object) tuple. 
~\cite{zhang2015feature} decomposes the task in the feature space as principle and non-principle features. ~\cite{devin2019compositional} reasons task decomposition from the object-centric way. 
The task decomposition considered in our work is most similar to~\cite{devin2017learning}, in which tasks are decomposed by a set of degrees of variation. 
By leveraging knowledge from different tasks with common degrees of variation, we aim to directly obtain policies for novel combinatorial tasks with existing degrees of variation. 
However, unlike~\cite{devin2017learning} that performs task decomposition through a modular policy architecture, we achieve task decomposition in the context of meta-RL by disentangling the latent task representation.

\section{Our Proposed Approach}
\label{sec:method}
In this section, we first give a brief review of prior works on context-based meta-RL algorithms in Section~\ref{sec:method:preliminary}. This provides the background to formalize our problem setup in Section~\ref{sec:method:formulate}. Afterwards, we describe our method in two parts: how to enable task decomposition in the latent task representation during training in Section~\ref{sec:method:training}, and zero-shot policy generalization leveraging the achieved decomposition in Section~\ref{sec:method:generalization}.


\subsection{Preliminary: Context-based Meta-RL}
\label{sec:method:preliminary}
Meta-RL typically assumes a distribution of tasks $p(\mathcal{T})$. Each task $\mathcal{T}$ is modeled as an independent Markov decision process (MDP), $\mathcal{M} \equiv (\mathcal{S}, \mathcal{A}, \mathcal{P}, R)$, where $\mathcal{S}$ corresponds to the state space, $\mathcal{A}$ corresponds to the action space, $\mathcal{P}$ denotes the  transition probability, and $R$ represents the real-valued reward. Meta-RL algorithms learn the task-conditioning policy from training tasks sampled from $p(\mathcal{T})$, and adapt the learned policy to new tasks.
Specifically, we focus on one perspective of meta-RL, \textit{i.e.} context-based meta-RL, which views meta-learning as task inference~\cite{rakelly2019efficient, zhao2020meld}. These methods, with an encoder $q_{\phi}$, map each task $\mathcal{T}_i$ into a task embedding $\mathbf{z}_i$ based on the past experience on this task, \textit{e.g.} context $\mathbf{c}_i=\{(\mathbf{s}_{t}, \mathbf{a}_{t}, \mathbf{s}_{t}^{\prime}, r_{t})\}_{t=1...N}$ \cite{rakelly2019efficient}. The policy $\pi_{\theta}$ is conditioned on the task embedding so that it can be adapted to different tasks, \textit{i.e.},
\begin{equation}
    \mathbf{a}\sim\pi_{\theta}(\cdot|\mathbf{s},\mathbf{z}_i),\quad \text{where}\ \mathbf{z}_i\sim q_{\phi}(\cdot|\mathbf{c}_i).
    \label{eq:meta-rl}
\end{equation}

While our method is agnostic to the specific context-based meta-RL algorithms, in our implementation, we build on the framework of probabilistic embeddings for actor-critic RL (PEARL)~\cite{rakelly2019efficient}, which optimizes the objective,
\begin{equation}
    \mathbb{E}_{\mathcal{T}_i}\left[\mathbb{E}_{\mathbf{z}_i \sim q_{\phi}\left(\mathbf{z}_i \mid \mathbf{c}_{i}\right)}\left[R(\mathcal{T}_i, \mathbf{z}_i)+\beta D_{\mathrm{KL}}\left(q_{\phi}\left(\mathbf{z}_i \mid \mathbf{c}_{i}\right) \| p(\mathbf{z}_i)\right)\right]\right],
\end{equation}
where $R(\mathcal{T}_i, \mathbf{z}_i)$ denotes the addition of actor and critic objectives as defined in~\cite{rakelly2019efficient}. We refer the readers~\cite{rakelly2019efficient} for more technical details.

\begin{figure}[htb!]
    \centering
    \includegraphics[width=0.4\linewidth]{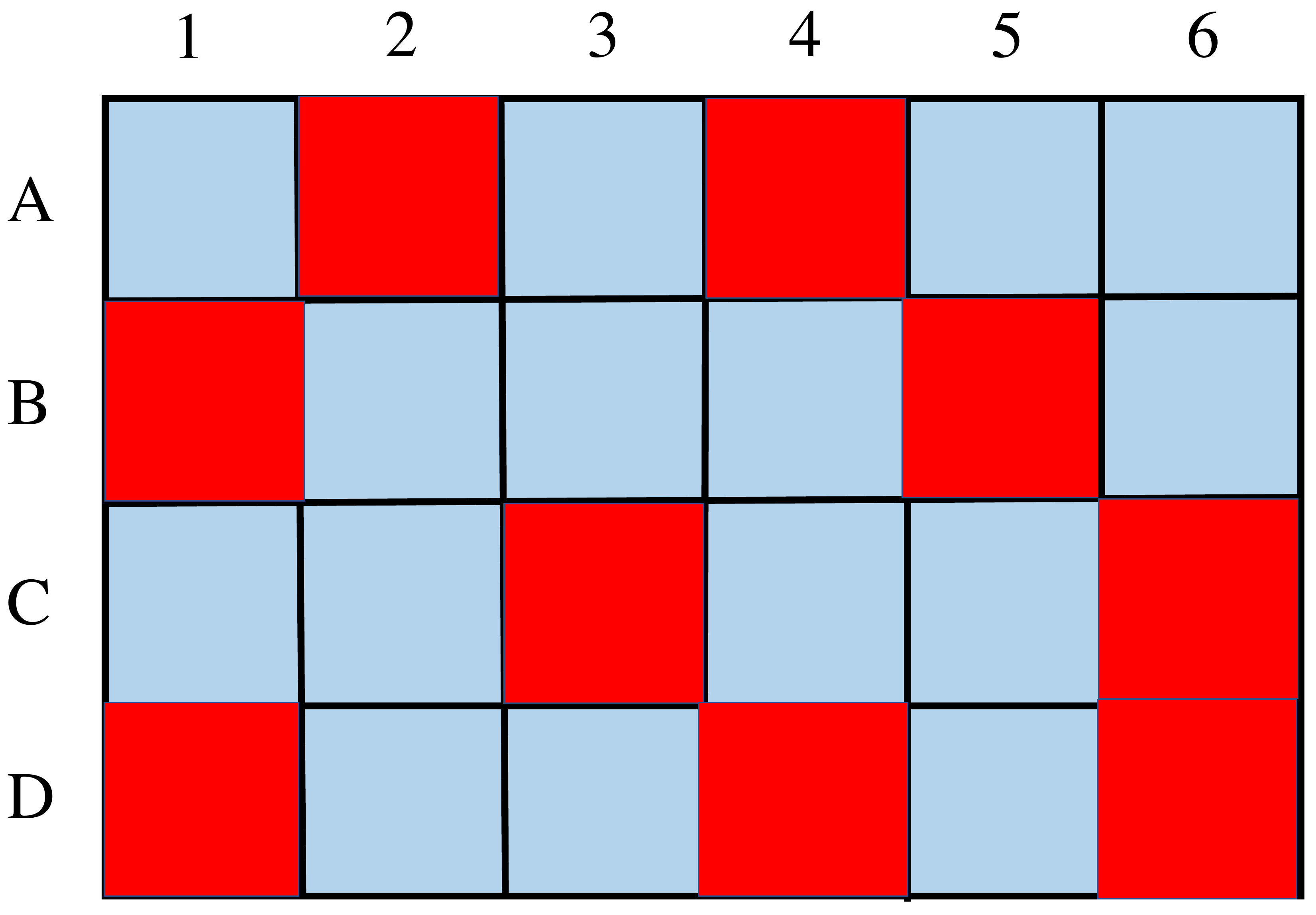}
    \caption{Illustration of the targeted policy generalization setting in our work. We assume the tasks-of-interest can be described as combinations of degrees of variation (DoV). Given experience from training task set $\mathbf{T}_{train}$ (blue), we aim to generalize across tasks (in red) whose DoV combinations are unseen but each DoV label is present in $\mathbf{T}_{train}$.}
    \label{fig:generalization_illustration}
\end{figure}

\subsection{Problem Formulation} 
\label{sec:method:formulate}
Our goal is to achieve zero-shot policy transfer across unseen compositional tasks. 
We define compositional task as the task equipped with a predefined set of degrees of variation (DoVs)~\cite{devin2017learning}. 
Loosely speaking, a task DoV describes one varied aspect of the task. 
For example, in Fig.~\ref{fig:motivation}(b), there are two DoVs describing the tasks, \textit{i.e.} \textit{environment} and \textit{goal}, and the DoV \textit{goal} has three unique values, \textit{i.e.} \textit{left} (red trajectory), \textit{front} (blue trajectory), and \textit{right} (yellow trajectory). Formally, considering the task distribution $p(\mathcal{T})$, each $\mathcal{T}_i$ is equipped with a unique combination $\mathcal{C}_i$ of $M$ task DoVs as $\mathcal{C}_i=(y_1^i,\dots, y_M^i)$, where $y_j^i$ is the label of the $j$-th DoV of task $\mathcal{T}_i$. The $M$ DoVs are problem-dependent and identical for all the tasks $\mathcal{T}_i$ in the task set $\mathbf{T}$, while each DoV can have different labels and the task DoV combination $\mathcal{C}_i$ for each task $\mathcal{T}_i$ is unique. 
We assume the DoV labels, $\mathcal{C}_i$, of each task $\mathcal{T}_i$ in $\mathbf{T}$ are given. 
Our problem setting aims at generalization across tasks whose DoV combinations are unseen but the DoV labels are present in training tasks $\mathbf{T}_{train}$, as illustrated in Figure~\ref{fig:generalization_illustration}. 

\subsection{Task Decomposition via Latent Task Embedding Disentanglement}
\label{sec:method:training}

\begin{figure*}[htb!]
    \centering
    \includegraphics[width=0.76\linewidth]{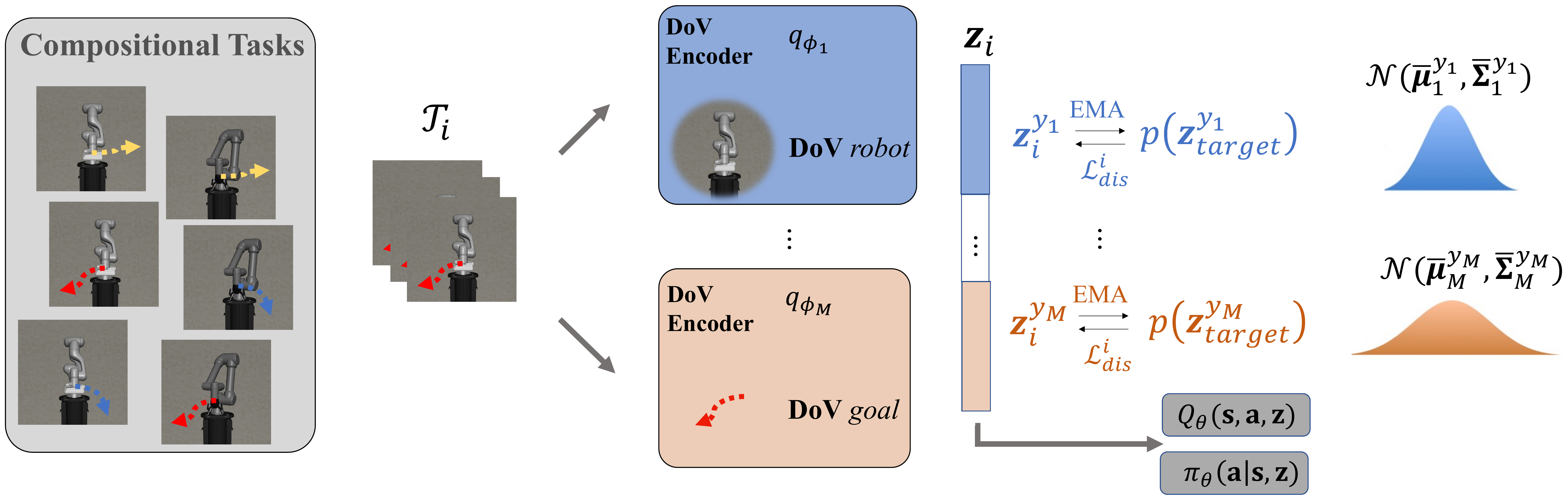}
    \caption{We achieve task decomposition via disentanglement in the latent task space. The disentanglement is achieved by encoding different task DoVs with individual DoV encoders and enforcing identical task DoVs across different tasks have similar embeddings.}
    \label{fig:model_overview}
\end{figure*}


In this section, we describe how we achieve task decomposition  \textit{w.r.t.} task DoV combinations in the context of meta-reinforcement learning (Meta-RL). Figure~\ref{fig:model_overview} illustrates our framework. The key insight is to achieve task decomposition by enforcing the disentanglement of the latent task representation in~\cite{rakelly2019efficient} during meta-training. While traditional meta-RL algorithms learn the task embedding $\mathbf{z}_i$ with a single encoder $q_{\phi}$ as Eq.~\ref{eq:meta-rl}, we replace it with $M$ separate DoV encoders $q_{\phi_1},q_{\phi_2},\dots,q_{\phi_M}$, where $q_{\phi_j}(\mathbf{z}_i^{y_j}|\mathbf{c}_i)$ learns the DoV embedding for the $j$-th DoV of task $\mathcal{T}_i$. The task embedding is obtained by concatenating all $M$ DoV embeddings as $\mathbf{z}_i=[\mathbf{z}_i^{y_1},\dots,\mathbf{z}_i^{y_M}]$, and the policy $\pi_{\theta}$ is conditioned on the disentangled task embedding:
\begin{equation}
    \begin{aligned}
    \mathbf{a}\sim\pi_{\theta}(\cdot|\mathbf{s},\mathbf{z}_i)\quad
    &\text{where}\ \mathbf{z}_i=[\mathbf{z}_i^{y_1},\dots,\mathbf{z}_i^{y_M}]\\
    &\mathbf{z}_i^{y_j}\sim q_{\phi_j}(\cdot|\mathbf{c}_i),j=1,2,\dots,M.
    \end{aligned}
    \label{eq:ours}
\end{equation}

For different tasks $\mathcal{T}_u,\mathcal{T}_v$ sharing the same label in the $j$-th DoV, \textit{i.e.}, $y_j^u=y_j^v=y_j$, their corresponding DoV embedding should be regularized to follow an identical probability distribution. One naive way to implement this principle is to bring close the corresponding DoV embedding of different tasks with shared DoV labels within the sampled data batch. However, we found empirically that this strategy is easily prone to the affection of noise, making the training process unstable. To this end, we introduce a target distribution $p(\mathbf{z}_{target}^{y_j})$ for the $j$-th task DoV with $y_j$ label, and minimize the Kullback–Leibler (KL) divergence between $q_{\phi_j}(\mathbf{z}_i^{y_j}|\mathbf{c}_i)$ and  $p(\mathbf{z}_{target}^{y_j})$. Inspired by~\cite{tarvainen2017mean,grill2020bootstrap}, Exponential Moving Average (EMA) is incorporated into our training scheme to update this target distribution during training, achieving robustness against outliers during training. For each DoV label $y_j$ of the $j$-th DoV, we maintain an individual target distribution $p(\mathbf{z}_{target}^{y_j})$. Following   PEARL~\cite{rakelly2019efficient}, our DoV embedding is modeled as Gaussian distribution. So we maintain the EMA for both the mean $\bm{\bar{\mu}}_{j}^{y_j}$ and variance $\bm{\bar{\Sigma}}_{j}^{y_j}$ of $p(\mathbf{z}_{target}^{y_j})$. This  task disentanglement regularization is written as follows for each task $\mathcal{T}_i$, where $\lambda$ is a hyper-parameter to balance different loss items.
\begin{equation}
    \begin{aligned}
        &\mathcal{L}_{dis}^i=\lambda\sum_{j=1}^M D_{KL}(q_{\phi_j}(\mathbf{z}_i^{y_j}|\mathbf{c}_i)||p(\mathbf{z}_{target}^{y_j}))\\ &p(\mathbf{z}_{target}^{y_j})=\mathcal{N}(\bm{\bar{\mu}}_{j}^{y_j},\bar{\bm{\Sigma}}_{j}^{y_j})
    \end{aligned}
    \label{eq:disentangle}
\end{equation}
where $q_{\phi_j}(\mathbf{z}_i^{y_j}|\mathbf{c}_i)=\mathcal{N}(\bm{\mu}_i^{y_j},\bm{\Sigma}_i^{y_j})$ is the current prediction for task $\mathcal{T}_i$. $\bm{\bar{\mu}}_{j}^{y_j}$ and $\bar{\Sigma}_{j}^{y_j}$ are updated in each training step $t$ with EMA:
\begin{equation}
    \begin{aligned}
        &\bm{\bar{\mu}}_{j}^{y_j,t}\leftarrow \tau\cdot \bm{\bar{\mu}}_{j}^{y_j,t-1} + (1-\tau)\cdot \bm{\mu}_i^{y_j}\\  &\bm{\bar{\Sigma}}_{j}^{y_j,t}\leftarrow \tau\cdot \bm{\bar{\Sigma}}_{j}^{y_j,t-1} + (1-\tau)\cdot \bm{\Sigma}_i^{y_j},
    \end{aligned}
    \label{eq:ema}
\end{equation}
where $\tau=0.99$ in our implementation. The task disentanglement regularization is applied during meta-training, as summarized in Algorithm~\ref{algo:meta-train}, where red lines highlight the differences compared with PEARL.

\begin{algorithm}[htb!]
\caption{Our meta-training procedure. Differences with PEARL are highlighted in \textcolor{red}{red}.}
\begin{algorithmic}[1]
    \REQUIRE Batches of compositional training tasks $\{\mathcal{T}_i\}_{i=1...T}$ \textcolor{red}{with corresponding task DoV combinations $\{\mathcal{C}_i\}_{i=1...T}$}, learning rate $\alpha_1, \alpha_2, \alpha_3$
    
    \STATE Init. replay buffer $\mathcal{B}_i$ for each training task
    \WHILE {not done}
    
        \FOR{each task $\mathcal{T}_i$}
            \STATE Initialize context $\mathbf{c}_i = \{\}$
            \FOR{$k=1,...,K$}
              \STATE \textcolor{red}{Sample $\mathbf{z}_i^{y_j} \sim q_{\phi_j}(\cdot|\mathbf{c}_i),j=1,\dots,M$}
              \STATE \textcolor{red}{Concatenate task embedding $\mathbf{z}_i=[\mathbf{z}_i^{y_1},\dots,\mathbf{z}_i^{y_M}]$}
              \STATE Gather data from $\pi_\theta(\mathbf{a} | \mathbf{s,z}_i)$ and add to $\mathcal{B}_i$ and update $\mathbf{c}_i = \{(\mathbf{s}_l, \mathbf{a}_l, \mathbf{s}_l^{\prime}, \mathbf{r}_l)\}_{l:1...N} \sim \mathcal{B}_i$
            \ENDFOR
        \ENDFOR
        
        \FOR{step in training steps}
            \FOR{each $\mathcal{T}_i$}
                \STATE Sample context batch $\mathbf{c}_i \sim \mathcal{S}_c(\mathcal{B}_i)$ and RL batch $b_i \sim \mathcal{B}_i$
                \STATE \textcolor{red}{Sample $\mathbf{z}_i^{y_j} \sim q_{\phi_j}(\cdot|\mathbf{c}_i),j=1,\dots,M$}
                \STATE \textcolor{red}{Concatenate task embedding $\mathbf{z}_i=[\mathbf{z}_i^{y_1},\dots,\mathbf{z}_i^{y_M}]$}
                \STATE Compute $\mathcal{L}_{actor}^{i},\mathcal{L}_{\text {critic }}^{i},\mathcal{L}_{KL}^{i}$ same as PEARL
                \STATE \textcolor{red}{$\mathcal{L}_{dis}^i=\lambda\sum_{j=1}^M D_{KL}(q_{\phi}(\mathbf{z}_i^{y_j}|\mathbf{c}_i)||p(\mathbf{z}_{target}^{y_j}))$, 
                where $p(\mathbf{z}_{target}^{y_j})=\mathcal{N}(\bm{\bar{\mu}}_{j}^{y_j},\bm{\bar{\Sigma}}_{i}^{y_j})$}
                \STATE  \textcolor{red}{$\bm{\bar{\mu}}_{j}^{y_j,t}\leftarrow \tau\cdot \bm{\bar{\mu}}_{j}^{y_j,t-1} + (1-\tau)\cdot \bm{\mu}_i^{y_j}$}
                \STATE
               \textcolor{red}{$\bm{\bar{\Sigma}}_{j}^{y_j,t}\leftarrow \tau\cdot \bm{\bar{\Sigma}}_{j}^{y_j,t-1} + (1-\tau)\cdot \bm{\Sigma}_i^{y_j}$}
            \ENDFOR
            \STATE \textcolor{red}{$\phi_j \leftarrow \phi_j-\alpha_{1} \nabla_{\phi_j} \sum_{i}\left(\mathcal{L}_{critic}^{i}+\mathcal{L}_{K L}^{i}+\mathcal{L}_{dis}^{i}\right),j=1,\dots,M$}
            \STATE Update $\theta_{\pi},\theta_{Q}$ same as PEARL
        \ENDFOR
    
    \ENDWHILE
\end{algorithmic}
\label{algo:meta-train}
\end{algorithm}

\subsection{Zero-Shot Policy Generalization through Task Decomposition}
\label{sec:method:generalization}
When generalizing policies to novel tasks, traditional meta-RL algorithms usually requires adequate explorations on the new tasks. However, exploration can be costly in some scenarios, \textit{e.g.}, the task objects are fragile such as the thru-hole components in PCB assembly.
In contrast, we can achieve zero-shot policy generalization leveraging the task decomposition obtained from Section~\ref{sec:method:training}.
We identify two scenarios of zero-shot policy generalization as follows.

\paragraph{S1} Test tasks are defined by combinations of DoV labels already seen in training tasks $\mathbf{T}_{train}$. In this case, we directly concatenate the EMA $\{\mathbf{z}_{target}^{y_j}\}_{j=1}^M$ of the corresponding DoV embeddings obtained from training as the test task representation. The meta-policy conditioned on the concatenated task representation is used as the generalized policy for the new task.

\paragraph{S2} Test tasks have DoV label that does not exist in the training tasks $\mathbf{T}_{train}$, denoted as, $y_j=y^*$. Similar to PEARL meta-test, our method first collects context in one single task $\mathcal{T}_{test}$ with the unseen DoV label and acquires the latent task representation $\mathbf{z}_{test}$. Then the embedding of the DoV label, $\mathbf{z}_{test}^{y_j}$, is inferred as the corresponding dimensions extracted from  $\mathbf{z}_{test}$. After obtaining the embedding of this unseen DoV label, we repeat the steps in \textbf{S1} to achieve generalization across other test tasks with $y_j=y^*$. While our method does require collecting experiences from one single task in test task set, the generalization across other test tasks is still \textbf{zero-shot}. As demonstrated in Section~\ref{sec:exp:real}, this can be readily used as a sim-to-real generalization method. For instance, we meta-train a policy to complete $10$ similar manipulation tasks IN different simulation environments. When generalizing to the real world, our method only requires interacting with the real environment on one single task, and is able to generalize to the other $9$ tasks in a zero-shot fashion. 

\section{Experimental Results}
\label{sec:exp}
In the experiments, we aim to investigate the following questions. 1) Does our method described in Section~\ref{sec:method:training} effectively disentangle the learned task representation? 
2) How does the disentanglement affect the model training for compositional tasks? 
3) Can our method achieve zero-shot policy generalization across novel tasks through task decomposition?

\subsection{Simulation Experiments}
\label{sec:exp:simulation}

\begin{figure}[htb!]
    \centering
    \includegraphics[width=\linewidth]{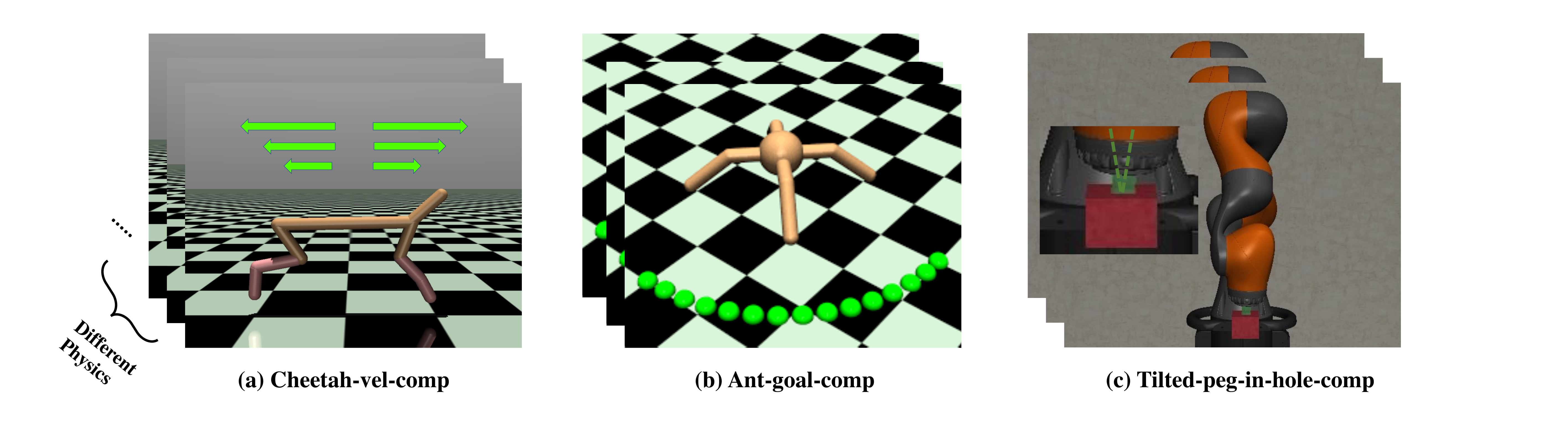}
    \caption{Three set of compositional tasks for algorithm evaluation. (a) Cheetah-vel-comp with varying \textit{goal velocities} and \textit{physics}. (b) Ant-goal-comp with varying \textit{goal positions} and \textit{physics}. (c) Tilted-peg-in-hole-comp with varying \textit{hole tilting directions} and \textit{physics}.} 
    \label{fig:exp_setup}
    \vspace{-3mm}
\end{figure}

\paragraph{Compositional Task Setup.}
To study the targeted policy transfer problem in this work, we build three set of simulated compositional tasks shown in Figure~\ref{fig:exp_setup}. Each set of compositional tasks has two task degrees of variation (DoVs), \textit{i.e.,} \textit{goal} and \textit{physics}. For (a) Cheetah-vel-comp and (b) Ant-goal-comp, we extend the traditional Cheetah-vel and Ant-goal from~\cite{finn2017model} by randomizing some physical parameters, \textit{i.e.}, \textit{body mass, body inertia, damping} and \textit{friction}. Specifically, we randomly sample $20$ different sets of physical parameters and $20$ different goal velocities for Cheetah-vel-comp, leading to $400$ compositional  tasks in total. Thus each task has a unique \textit{(physics, goal)} DoV combination. Similarly for Ant-goal-comp, $15$ different groups of physical parameters and $15$ different goal 2D locations are randomly sampled. 
In (c) Tilted-peg-in-hole-comp, we create challenging tilted peg-in-hole tasks based on robosuite~\cite{robosuite2020}. The tasks are performed using a 7-Degree-of-Freedom (DoF) KUKA LBR IIWA robot with operation space control~\cite{khatib1987unified}. We adopt a square peg-hole task with $0.5 mm$ clearance and tilted the hole surface plane $5^{\circ}$ towards different directions. Tilted peg-in-hole tasks are prone to jamming issues during execution, thus posing more challenges to robust policy learning. The observation space is $6$-dimensional end-effector pose and the actions are the desired end-effector poses. The compositional tasks are composed of $4$ different hole tilting directions towards $\pm x$ and $\pm y$ axis separately as well as $20$ different sets of physical parameters (\textit{i.e. friction between peg and hole, damping} and \textit{controller step size}). 


\paragraph{Training Tasks Selection.}
The training task set $\mathbf{T}_{train}$ is a subset of the whole compositional task set $\mathbf{T}$. When selecting $\mathbf{T}_{train}$, we 
randomly select tasks in $\mathbf{T}$ with probability $\alpha\in(0,1)$, while ensuring each task DoV label in $\mathbf{T}$ appear at least once in $\mathbf{T}_{train}$. For example, in Figure~\ref{fig:generalization_illustration}, the 15 tasks in blue are selected for training ($\alpha=0.625$). In the implementation, we use $\alpha = 0.5$, which is discussed in Section~\ref{sec:exp:ablation}. Other tasks in $\mathbf{T}$ are set aside as novel test tasks, \textit{i.e.} $\mathbf{T}_{test}=\mathbf{T}\backslash \mathbf{T}_{train}$.

\begin{figure}[htb!]
    \centering
    \includegraphics[width=\linewidth]{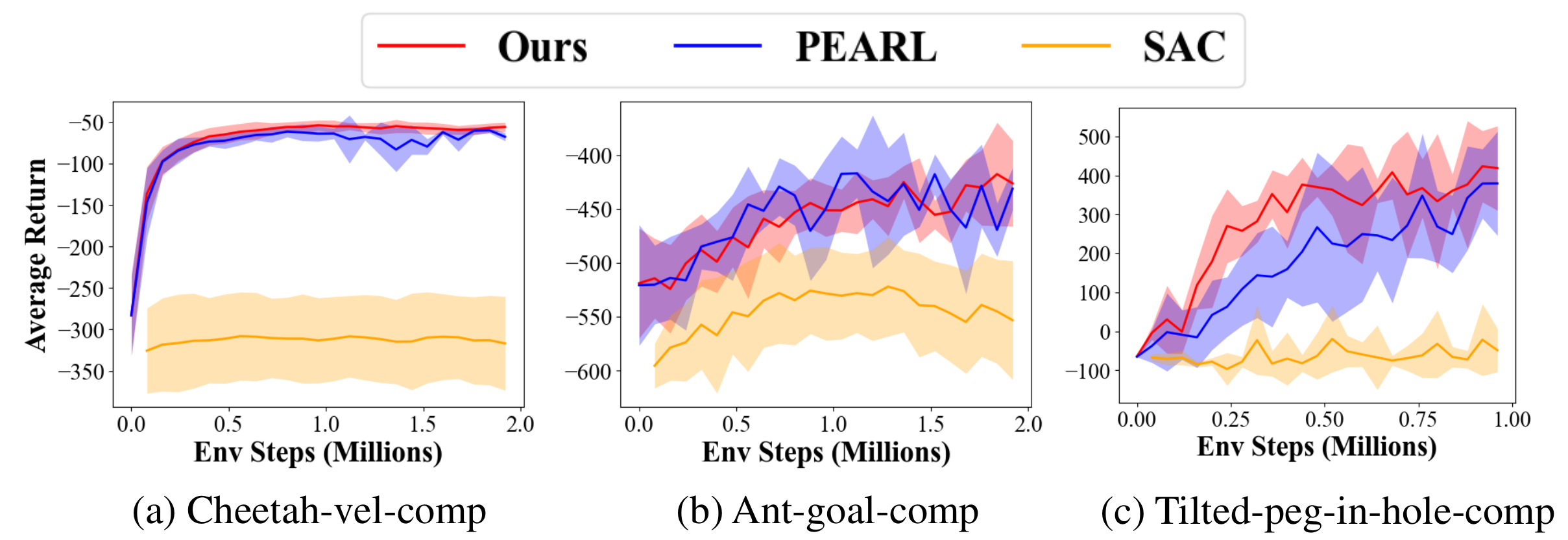}
    
     \caption{Average return on test tasks during the training period. Our method and PEARL apply the same meta-test strategy, where the first two trajectories are aggregated into the context for task inference. } 
    \label{fig:result:progress} 
\end{figure}

\paragraph{Baselines.} We compare our method against two baselines, PEARL~\cite{rakelly2019efficient} and SAC~\cite{haarnoja2018soft}. PEARL directly learns the task embedding from training tasks without exploiting the inherent task combinatorial structure. During generalization, \textit{i.e.}, meta-test, PEARL requires extra explorations on each test task, so it is not a zero-shot approach.
The SAC agent is trained with all the training tasks together where a training task is randomly sampled in each episode. This can be viewed as a variant of domain randomization. 
The learned SAC policy is directly applied to test tasks. 
It is worth noting that~\cite{devin2017learning} can not be directly applied as a baseline because solely decomposing tasks \textit{w.r.t.} observations in the three tasks fail to capture the dynamics differences, as discussed in Section~\ref{sec:inrto}.

\paragraph{Effect on Training.} 
Figure~\ref{fig:result:progress} shows the performance of the learned policies on $\mathbf{T}_{test}$ as training proceeds, across $5$ random seeds. Our method achieves comparable meta-test results compared to PEARL, indicating that the enforced regularization in latent task representation (Section~\ref{sec:method:training}) does not deteriorate the learning process of meta-RL. It is worth noting that the aim of our proposed method is to improve the performance of zero-shot policy transfer across test tasks, instead of meta-test where the experience of the test tasks is given.

\paragraph{Zero-Shot Evaluation.}

\begin{figure*}[htb!]
\centering
\begin{minipage}[t]{0.72\linewidth}
    \centering
    \includegraphics[width=\linewidth]{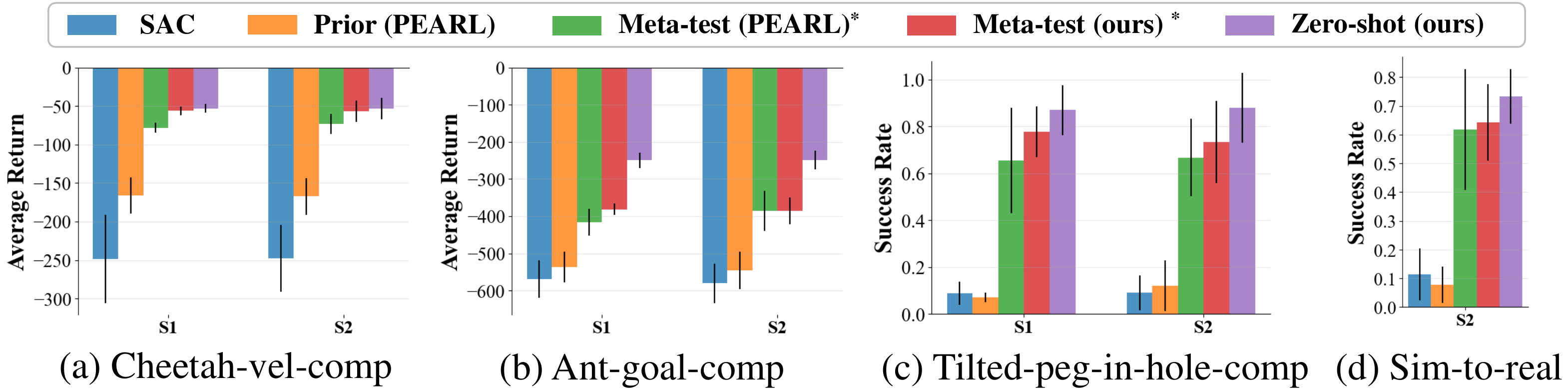}
    \caption{Evaluation results of policy generalization on test tasks ($^*$ not a zero-shot method).} 
    \label{fig:result:zero-shot}
\end{minipage}
\hspace{3mm}
\begin{minipage}[t]{0.25\linewidth}
    \centering
    \includegraphics[width=\linewidth]{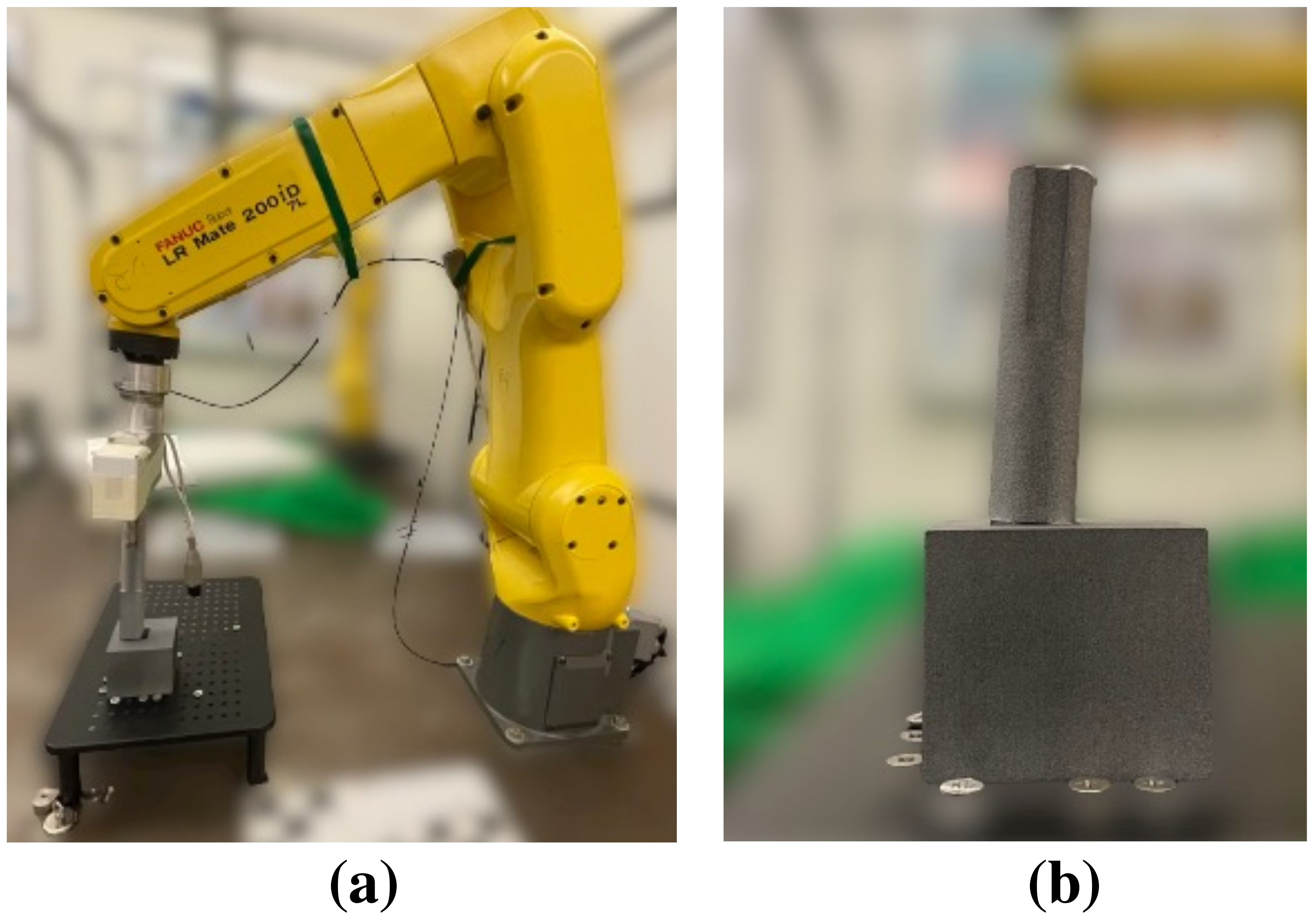}
    \caption{(a) Real-world setup. (b) $5 ^{\circ}$ tilted hole for experiment.} 
    \label{fig:real_setup}
\end{minipage}
\end{figure*}

The evaluation is conducted for the two scenarios \textbf{S1} and \textbf{S2} (Section~\ref{sec:method:generalization}) respectively. 
In \textbf{S1}, $\mathbf{T}_{test}=\mathbf{T}\backslash \mathbf{T}_{train}$ is used as the test tasks. 
While in \textbf{S2}, we first sample task attributes that do not exist in $\mathbf{T}$ and the test tasks are set as the combinations of the unseen attributes and other seen attributes, e.g., (\textit{unseen} physics, \textit{seen} goal), (\textit{unseen} goal, \textit{seen} physics). 
The experimental results across 5 random seeds are shown in Figure~\ref{fig:result:zero-shot}. We compare our zero-shot performance (\textit{Zero-shot (ours)}) with multiple baselines: 
1) \textit{SAC:} A single model is trained on all training tasks with  SAC  \cite{haarnoja2018composable}.
2) \textit{Prior (PEARL):} The same $\mathcal{N}(0,1)$ prior distribution is applied to all the task embeddings $\mathbf{z}_i$ of PEARL model \cite{rakelly2019efficient}. 
3) \textit{Meta-test (PEARL):} We apply the same meta-test for PEARL as in \cite{rakelly2019efficient}. To ensure enough contexts on the test task, the first 8 trajectories are aggregated into the context. 
4) \textit{Meta-test (ours):} The same meta-test with PEARL is applied to the model trained with our method. 
It is worth mentioning that only 1) and 2) can achieve zero-shot generalization on novel tasks, while 3) and 4) require extra exploration. As shown in Figure~\ref{fig:result:zero-shot}, our zero-shot generalization significantly outperforms the two zero-shot baselines, \textit{SAC} and \textit{Prior (PEARL)}. Interestingly, it is also superior to the two meta-test baselines, although our generalization does not explore on the test task. We hypothesize that the reason is our method benefits from the EMA embedding $\mathbf{z}_{target}^{y_j}$ used to compose the embedding $\mathbf{z}_u$ of the novel task, serving as a temporal ensemble of information accumulated during training. In contrast, purely collecting context from a single test task may suffer from its inherent bias and easily get trapped in the local optimum.

\begin{figure}[htb!]
    \centering
    \includegraphics[width=\linewidth]{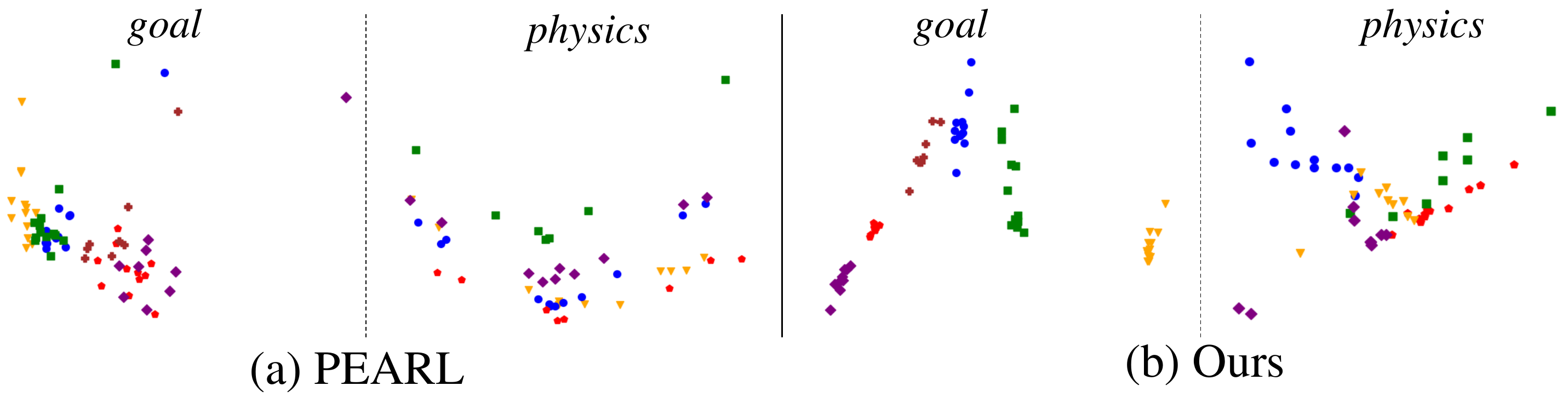}
    \caption{Visualization of latent task embedding of our method compared to PEARL. The embeddings of tasks with identical \textit{goal} or \textit{physics} labels are in the same color.} 
    \label{fig:latent_visualizations}
\end{figure}

\paragraph{Visualization of Latent Embedding.} To investigate if the proposed method in Section~\ref{sec:method:training} effectively disentangle the encodings in latent space, Figure~\ref{fig:latent_visualizations} visualizes sampled test task representations output by our method and PEARL on Cheetah-vel-comp. The embeddings of test tasks with same \textit{goal} or \textit{physics} are in the same color. 
All the embeddings are reduced to 2-d with PCA~\cite{abdi2010principal} for visualization. It is observed that our method can distinguish different labels of each attribute better than PEARL. Different goals are distinctly separable by our method. For different physics, despite their effect being indirect, similar physics parameters still tend to induce closer embeddings.

\subsection{Real Experiments}
\label{sec:exp:real}
As described in Section~\ref{sec:method:generalization}, our method can be applied as a sim-to-real algorithm, by treating real-world tasks as tasks with unseen \textit{physics} attribute. We evaluate the method on a real-world 6-DoF FANUC Mate 200iD robot with the similar setup as Tilted-peg-in-hole-comp in Section~\ref{sec:exp:simulation}, as shown in Figure~\ref{fig:real_setup}.
The evaluation of real world is identical to the evaluation of \textbf{S2} in simulation (Section~\ref{sec:exp:simulation}), where, after meta-training in simulation, the meta-policy collects context for one held-out task to infer the real-world physics embedding. 
The obtained physics embedding is combined with other goal embeddings learned from simulation to generalize across the other three tasks in the real world. Each task is used as the held-out task once for thorough evaluation. When evaluating the generalized policy, the evaluated task is executed $30$ trials with $3$ different random seeds ($10$ trials for each random seed). The results are reported in Figure~\ref{fig:result:zero-shot}(d). Our generalized policy achieves a success rate of $73.3\%$ over different unseen tasks. 

\begin{figure}[htb!]
    \centering
    \begin{subfigure}{0.48\linewidth}
         \centering
         \includegraphics[width=\textwidth]{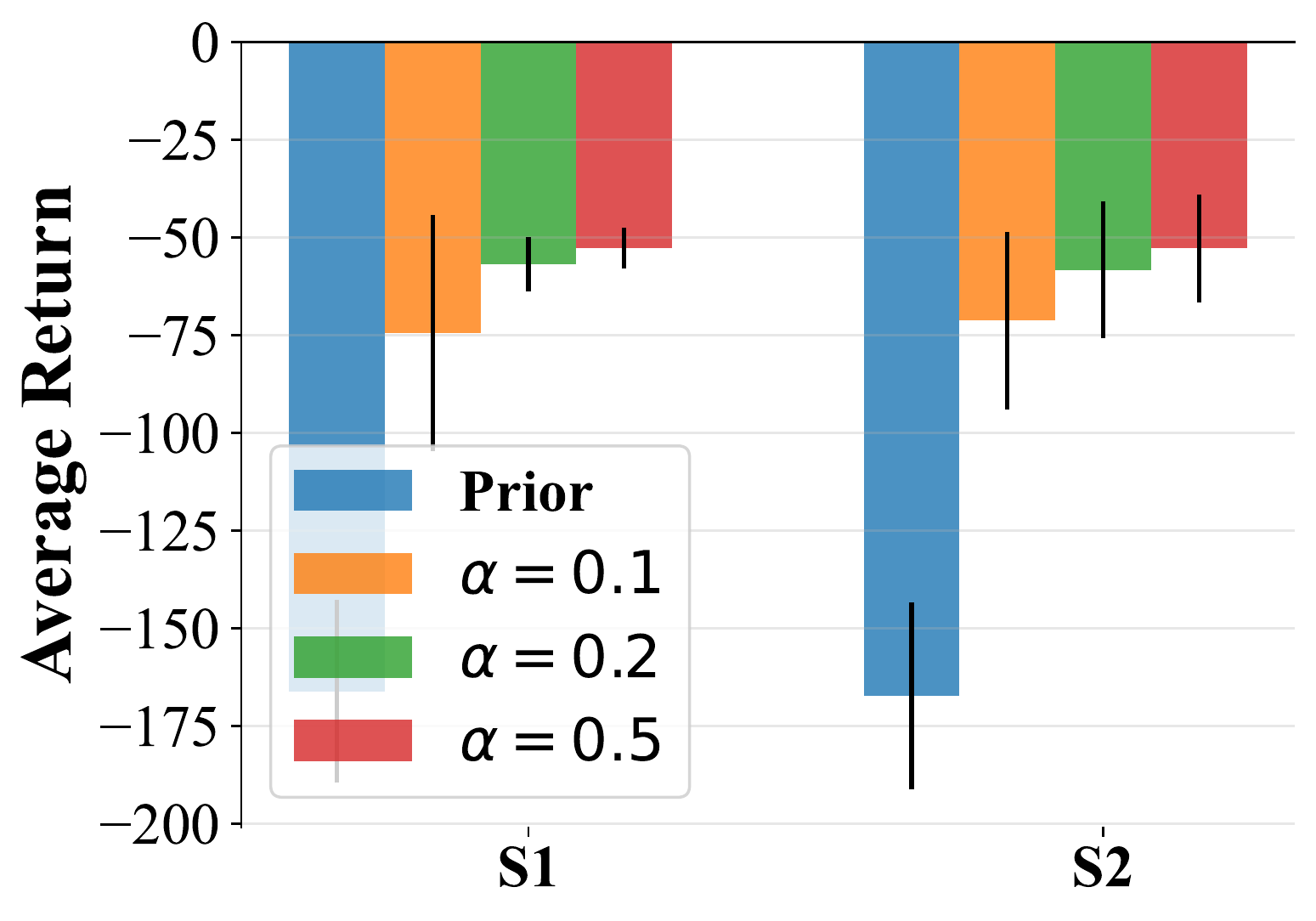}
         \caption{Training Task Sparsity}
        \label{fig:result:ablation1}
     \end{subfigure}
    \begin{subfigure}{0.48\linewidth}
         \centering
         \includegraphics[width=\textwidth]{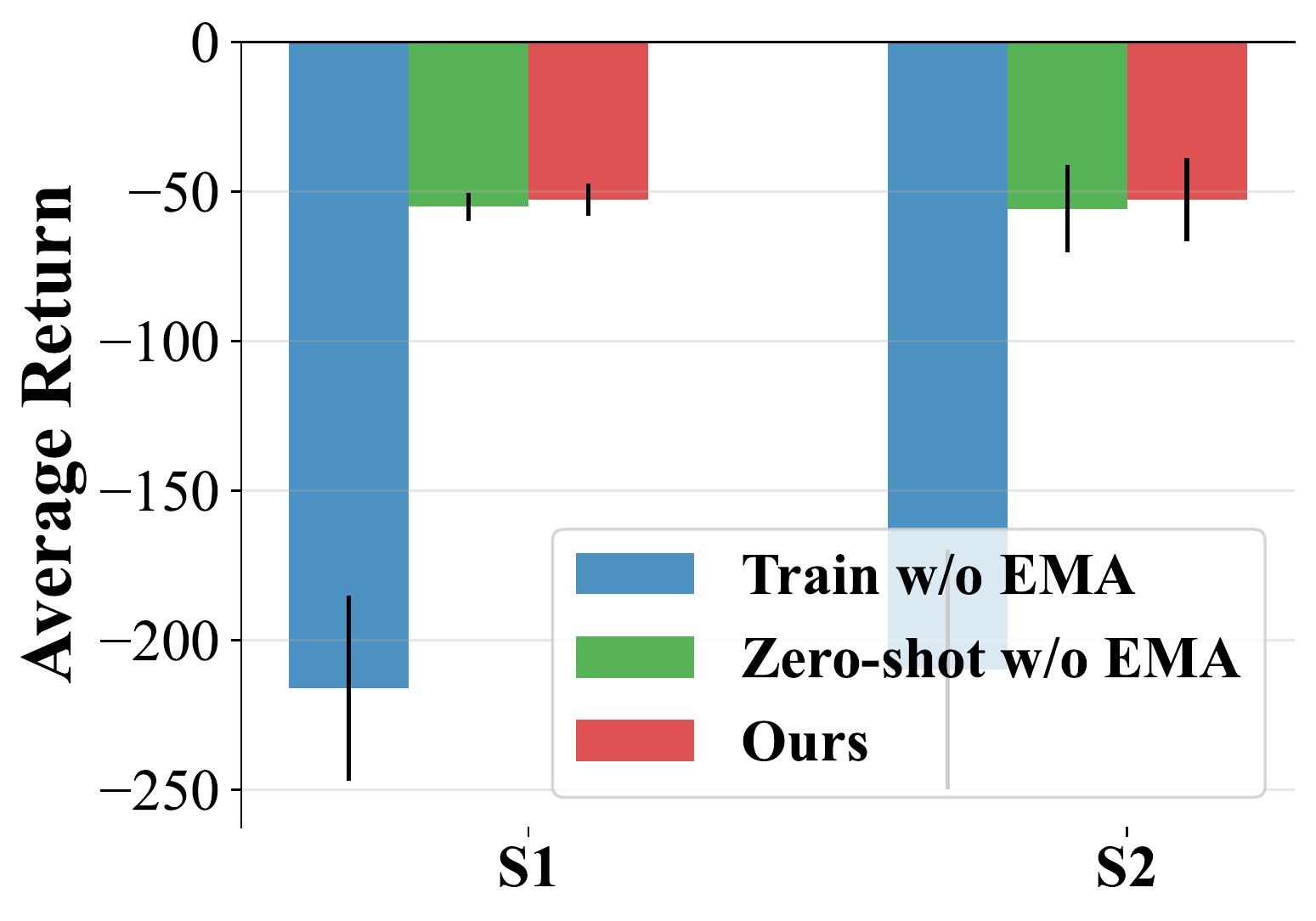}
         \caption{Effect of EMA}
        \label{fig:result:ablation2}
    \end{subfigure}
    \caption{Results of ablation studies.}
    \label{fig:result:ablation}
\end{figure}

\vspace{-5mm}
\subsection{Ablation Studies}
\label{sec:exp:ablation}

\paragraph{Sparsity of Training Tasks.}
In our previous experiments, we sample 50\% of compositional tasks (\textit{i.e.} $\alpha=0.5$, see Section~\ref{sec:exp:simulation} (b) for details) for training, so that each task DoV label would appear multiple times during training. To examine the limit of our method, we try to reduce the number of training tasks, \textit{i.e.} $\alpha=0.2$ or $0.1$, on Cheetah-vel-comp. In this case, each task DoV label would appear only $4$ or $2$ times on average among training tasks. Our zero-shot generalization performance is visualized in Figure~\ref{fig:result:ablation1}. Despite the performance decrease with fewer training tasks, our approach still notably outperforms the \textit{Prior} even with the fewest training tasks ($\alpha=0.1$). This indicates  our method learns meaningful disentangled DoV embeddings even with sparse training tasks.

\paragraph{Effect of EMA.} To justify the role of Exponential Moving Average (EMA) in our training and zero-shot generalization, we consider the following alternative strategies: 1) \textit{training w/o EMA:} Equivalently, we set $\tau=0$ in Eq.~\ref{eq:ema}. In this case, the task DoV embedding is regularized by the training task with the same DoV label sampled in the last training batch. 2) \textit{zero-shot generalization w/o EMA:} We try another strategy to get the corresponding DoV embeddings for an unseen task $\mathcal{T}_u$ instead of using EMA in \textbf{S1} of Section~\ref{sec:method:generalization}. The average DoV embeddings from training tasks with the same task DoV label with $\mathcal{T}_u$ is adopted instead, \textit{i.e.} $\mathbf{z}_u^{y_j}=\textbf{\textrm{mean}}\left(\{\mathbf{z}_v^{y_j}|\mathcal{T}_v\in\mathbf{T}_{train},y_j^u=y_j^v\}\right)$. 
The results are shown in Figure~\ref{fig:result:ablation2}. Training without EMA fails because of the large noise in the regularization of the DoV embeddings. EMA ``smoothes out'' the noise, serving as a temporal ensemble of context accumulated during training. In addition, it provides unbiased DoV embeddings for zero-shot generalization, as indicated by the generalization performance gap with and without EMA.

\section{Conclusion and Future Work}

In this paper, we propose a meta-RL algorithm to achieve zero-shot policy generalization across unseen compositional tasks. Our method learns disentangled task representation by encoding each task DoV separately. The DoV embeddings obtained from training tasks enable zero-shot policy generalization on test tasks by composing the task embeddings of novel tasks. Our approach achieves significantly superior performance on simulation tasks and can be seamlessly applied to sim-to-real generalization. 

The experiments conducted in this work are mainly the combination of two task DoVs, \textit{i.e.} \textit{physics} and \textit{goal}. In future work, we hope to extend to more diverse attributes, allowing for wider applications of zero-shot policy generalization.  Besides, one implicit assumption of modeling task decomposition via disentanglement is the specified task DoVs are independent of each other, limiting the application scenarios. Discovering other ways to achieve task decomposition without this assumption is another interesting direction to explore.




\addtolength{\textheight}{-2cm}   









\bibliographystyle{IEEEtran}
\bibliography{references}

\end{document}